\documentclass[10pt,twocolumn,letterpaper]{article}

\usepackage{iccv}
\usepackage{times}
\usepackage{epsfig}
\usepackage{graphicx}
\usepackage{amsmath}
\usepackage{amssymb}
\usepackage{makeidx}  
\usepackage{multirow}
\usepackage[pagebackref=true,breaklinks=true,letterpaper=true,colorlinks,bookmarks=false]{hyperref}

\iccvfinalcopy 

\def\iccvPaperID{3597} 

\begin{document}

\title{CoopSubNet: Cooperating Subnetwork for Data-Driven Regularization of Deep Networks under Limited Training Budgets}


\author{Riddhish Bhalodia$^1$\\
{\tt\small riddhishb@sci.utah.edu} \and 
Shireen Elhabian$^{1,2}$\\
{\tt\small shireen@sci.utah.edu} \and 
Ladislav Kavan$^2$\\
{\tt\small ladislav@cs.utah.edu} \and
Ross Whitaker$^{1,2}$\\
{\tt\small whitaker@cs.utah.edu} \\ \\
$^1$ Scientific Computing and Imaging Institute, University of Utah\\
$^2$ School of Computing, University of Utah }
\maketitle

\begin{abstract}
  Deep networks are an integral part of the current machine learning paradigm.   Their inherent ability to learn complex functional mappings between data and various target variables, while discovering hidden, task-driven  features, makes them a powerful technology in a wide variety of applications.
 Nonetheless, the success of these networks typically relies on the availability of sufficient training data to optimize a large number of free parameters while avoiding overfitting, especially for networks with large capacity.
 In scenarios with limited training budgets, e.g., supervised tasks with limited labeled samples, several generic and/or task-specific regularization techniques, including data augmentation, have been applied to improve the generalization of deep networks.
Typically such regularizations are introduced independently of that data or training scenario, and must therefore be tuned, tested, and modified to meet the needs of a particular network.  
  In this paper, we propose a novel regularization framework that is driven by the population-level statistics of the feature space to be learned. 
  The regularization is in the form of a \textbf{cooperating subnetwork}, which is an auto-encoder architecture attached to the feature space and trained in conjunction with the primary network.  We introduce the architecture and training methodology and demonstrate the effectiveness of the proposed cooperative network-based regularization in a variety of tasks and architectures from the literature. Our code is freely available at \url{https://github.com/riddhishb/CoopSubNet}.
\end{abstract}


\section{Introduction}
\label{sec:intro}

Deep neural networks (NNs) have shown to be an exceptionally powerful tool for machine learning in a variety of domains and applications.  The power of deep NNs is their ability to model complex \textit{functional mappings} between data and various target variables,
while discovering or generating features that facilitate this task.   This makes such networks an enabling technology in various tasks, including image classification \cite{he2016deep,chan2015pcanet}, pose estimation \cite{toshev2014deeppose,ranjan2019hyperface}, object detection \cite{ren2015faster,liu2016dhsnet}, face alignment \cite{ranjan2019hyperface,zhang2014facial,zhang2016learning}, and tracking \cite{nam2016learning,valmadre2017end}.

Deep networks are {\em data-hungry} models, and their performance relies heavily on the availability of large sets of training examples to learn complex relationships. This requirement for huge sets of observed examples stems from the large number of free parameters of these networks, that are typically necessary to capture complex relationships between often high-dimensional inputs 
(e.g., image textures and object geometries) and task-specific characteristics (e.g., classes of objects, positions, or pose).  Thus, state-of-the-art networks typically have parameters in the order of millions. For example, the VGGNet \cite{Simonyan15}, which identifies 1000 images categories, has 144 million parameters and requires 1.3 million training images with assigned categories (i.e., \textit{labeled data} in a supervised setting). 
%
Nonetheless, having millions of data samples, especially for supervised tasks where manual labeling and annotation are needed, is not always a viable option, especially for applications that entail high dimensional data (e.g., high resolution images, volumetric images) and/or data associated with significant monetary/computational costs (e.g., remote sensing, physical simulations). Ideally, deep networks would learn functional mappings from the data space to some unknown task-driven feature space while being generalizable to unseen samples, even with limited {\em training budget}.  
That is, the network should ideally not overfit to the limited training data, and their performance should degrade gracefully or marginally when an ideal training data set is not available.  
%

Regression problems (e.g., landmark detection \cite{ranjan2019hyperface,zhang2014facial,zhang2016learning}, image segmentation \cite{badrinarayanan2017segnet,chen2018deeplab}) are more prone to labeled data scarcity because obtaining large-enough training data for regression tasks is more challenging compared to classification tasks. 
One very effective approach for overcoming the challenge of overfitting with limited training data is to augment the training set.  
Data augmentation is the process of generating new data samples using the available limited data and other auxiliary information and/or assumptions about the domain, such as natural or expected invariants. In certain applications, however, data cannot be trivially augmented (e.g., regression) and is defined by a particular acquisition process that would be hard to synthesize (e.g., medical and astronomical images).
%
Furthermore, many applications rely heavily on data accuracy (e.g., tumor detection \cite{cruz2017accurate,shi2016stacked,sirinukunwattana2016locality}), which may make it challenging to augment data while ensuring accuracy.    Finally, the proposed method for regularization can be used to complement data augmentation strategies, and thus may be effective with or without these other methods.  
%

In conventional optimization problems and solution spaces, regularization is a common strategy for ill-posed problems, where it serves to restrict or prioritize the admissible solution space.
One way to regularize the solution space is to use a penalty on some property that may be expected from the solution a-priori (e.g., weight decay \cite{lang1990dimensionality}, smoothness of the learned mapping \cite{bengio2013representation}).
%
Such regularization methods are generic as they do not adapt to the task or the data at hand, and they often fail to capture population-level properties in the data and/or feature spaces that may violote, to some degree, the a-priori assumptions. 
%
%
Data-driven regularizers incorporate population statistics of the input data or the feature space and some very general assumptions about the complexity of those statistics, and they have been shown to outperform generic regularization methods, for instance using adaptive prior in Adaboost \cite{sriraam2013arboost}, for semi-supervised labeling task \cite{niyogi2006manifoldreg}, for classification \cite{pontial2005multipletask}, and for limited data training \cite{Zhu2018LDMNet}.
%
%
In particular, data/features in high-dimensional spaces tend to concentrate in the vicinity of a \textit{latent} manifold of a much lower dimensionality \cite{bengio2013representation}, which can be learned via principal component analysis (PCA) \cite{tipping1999probabilistic} or more recently using auto-encoders \cite{vincent2010stacked,bengio2013representation}. 

In this paper, we extend these previous observations on data-driven regularization and data modeling to deep neural networks, and propose a data-driven approach for learning network regularization that encodes population-level properties, as opposed to generic regularizers. 
%
Specifically, this paper systematically examines and analyzes the use of the latent \textit{features manifold} to regularize a deep network. 
%
To this regard, we propose a novel architecture of two interacting sub-networks, a \emph{primary network} that performs the main task (e.g., classification, regression), and a secondary \emph{cooperating subnetwork} (CoopSubNet) that is an auto-encoder attached to one of the intermediate feature spaces of the primary network. 
These sub-networks acts \emph{cooperatively} toward the given task. Being an auto-encoder, the secondary network learns a low-dimensional representation of the feature space, 
and cooperates with the primary network to enforce that the intermediate features adhere to a low-dimensional manifold. 

\section{Related Work}
\label{sec:related-works}
Limited data or low resource training models -- the focus of this paper -- is a crucial drawback of the current deep learning paradigm. Deep neural models are inherently data-hungry and, when faced with limited training budget they are prone to \emph{overfit} the data and generalize poorly. Literature has proposed several regularization models, network variants, and data augmentation schemes. A comprehensive literature review is beyond the scope of this paper. Here, we focus on the most closely related and relevant research to the proposed method.

The umbrella term {\em regularization} in the context of deep neural networks covers a wide range of different approaches to alleviate overfitting \cite{Kukacka2017RegularizationFD}. The class of methods dealing with penalty based regularization is of interest. This class includes classical approaches such as weight decay \cite{krogh1992simple}, and weight smoothing \cite{lang1990dimensionality}. There have been several different regularization terms, e.g., \cite{Rifai2011ContractiveAE, Sajjadi2016RegularizationWS}, but weight decay/$\mathbb{L}_2$ regularization on network weights remains the most widely used \cite{Kukacka2017RegularizationFD}. Most of these methods do not rely on data and hence have no information regarding the population variability of the features or the data. This allows them to be malleable to a variety of architectures and problems, but they fail to utilize the data/feature structure and statistics to improve generalization.

Another class of regularization methods modifies the training data (data augmentation). These modifications include adding noise to the input or features \cite{An1996Noise, Devries2017DatasetAI}, as well as image transformations \cite{Simard2003best}. Data augmentation usually relies on some modification to input data or features that is meant to capture invariant properties of the application (e.g., transformations that do not change the classification output).  This requires some knowledge of the domain/application, and it tends to work well in many classification tasks, where invariants are easier to formulate/imagine. Nonetheless, augmentation becomes nontrivial, if not infeasible, when we consider regression problems where invariants may be harder to discover. Data augmentation techniques are prevalent in the literature and in practice, and have proven to be very effective where applicable \cite{ekin2018autoaugment}. However, in domains where the data acquisition is expensive and the data robustness is critical (medical \cite{Razzak2018DLmedical} and astronomy \cite{kim2016galaxy}), generic data augmentation methods may fail and there is a need to devise data/task specific tactics to achieve effective data augmentation \cite{fridadar2018GANmedicalaug, bhalodia2018deepssm}.   Furthermore, even when data augmentation is feasible, its effectiveness may be enhanced by the application of some complementary regularization strategy. 

Network-architecture-driven regularization is another class of regularization methods, and this paper proposes a network-based framework to achieve data-driven regularization. Such regularization includes methods that use randomness in architecture (e.g., Dropout \cite{srivastava2014dropout} and Dropconnect \cite{Wan2013dropconnect}), modify layer operations (e.g., pooling and maxout units \cite{Goodfellow2013maxout}), and apply architectural modification (e.g., residual nets \cite{he2016deep} and skip-connections \cite{Long2015semantic}). All of these methods rely on data independent metrics to improve generalization and ignore population-level variations in the data and feature space of the given primary task.

Some of the earlier work hint at low-dimensional representation of the features in deep learning \cite{Rifai2011ContractiveAE}, and a recent method, \emph{LDMNet} \cite{Zhu2018LDMNet}, explores a similar idea of low-dimensional manifold based regularization. Nonetheless, LDMNet focuses on geometry driven representation of the joint manifold of features and data spaces that is specifically constructed for images. It further relies on the assumption of data coming from a set of low-dimensional manifolds. These assumptions limit the application of LDMNet to image classification problems, whereas we do not make any such assumptions about the data and the proposed regularization framework can be applied to any network architecture and learning task.

Another relevant advance is the advent of interacting networks - where one network is the {\em adversary} of the other, and the optimization seeks a solution to a min-max problem, as in generative adversarial networks (GANs) \cite{goodfellow2014GAN}. Some domain adaptation methods \cite{ganin2016domain} also use interacting networks that ensure the consistency of latent features across data domains.  Likewise, the TL-network trains two networks with different data inputs to share a latent space \cite{girdhar2016TLnet}.

In this paper, we propose a \emph{cooperating subnetwork} (CoopSubNet) that is trained in conjunction with the primary neural network. This subnetwork interacts in an unsupervised manner to enforce a penalty, or \textit{soft constraint}, on the primary network's intermediate features, which ensures they lie on or near a low-dimensional latent manifold. The reconstruction loss of the CoopSubNet is minimized along with the primary loss function, and hence the networks are \emph{cooperating}. The CoopSubNet provides a generic framework for a data-driven regularization for applications entailing limited training budget.

\section{Methods}
\label{sec:methods}

This paper deals with regularization methods that restrict a network or place penalties on the weights of a network in order to curtail overfitting and improve generalization.    
The hypothesis behind this work is that directly restricting the degrees of freedom in a network (e.g. bottleneck) or imposing penalties directly on the weights of a network not only avoid overfitting but also interfere with the ability of the network to learn or converge to an effective solution.    
As an alternative, we propose a \emph{cooperating subnetwork} (CoopSubNet) as a general, data-driven, regularization method for training deep networks under limited training data budgets. A CoopSubNet is attached to the primary network to introduce \textit{a soft penalty} that encourages the internal features of the primary network to lie close to a low-dimensional manifold.   A CoopSubNet can regularize a variety of primary network architectures and tasks and is added only during the training phase and therefore does not introduce any computational overhead during inference or testing. 
%
In this section, we start by introducing the CoopSubNet formulation, followed by the network training procedure. In the results section, we demonstrate the use of CoopSubNet applied to different tasks and network architectures and discuss alternative approach to cooperative networks (e.g., hard as opposed to soft constraints).


\begin{figure}[!h]
    \centering
    \includegraphics[width=\linewidth]{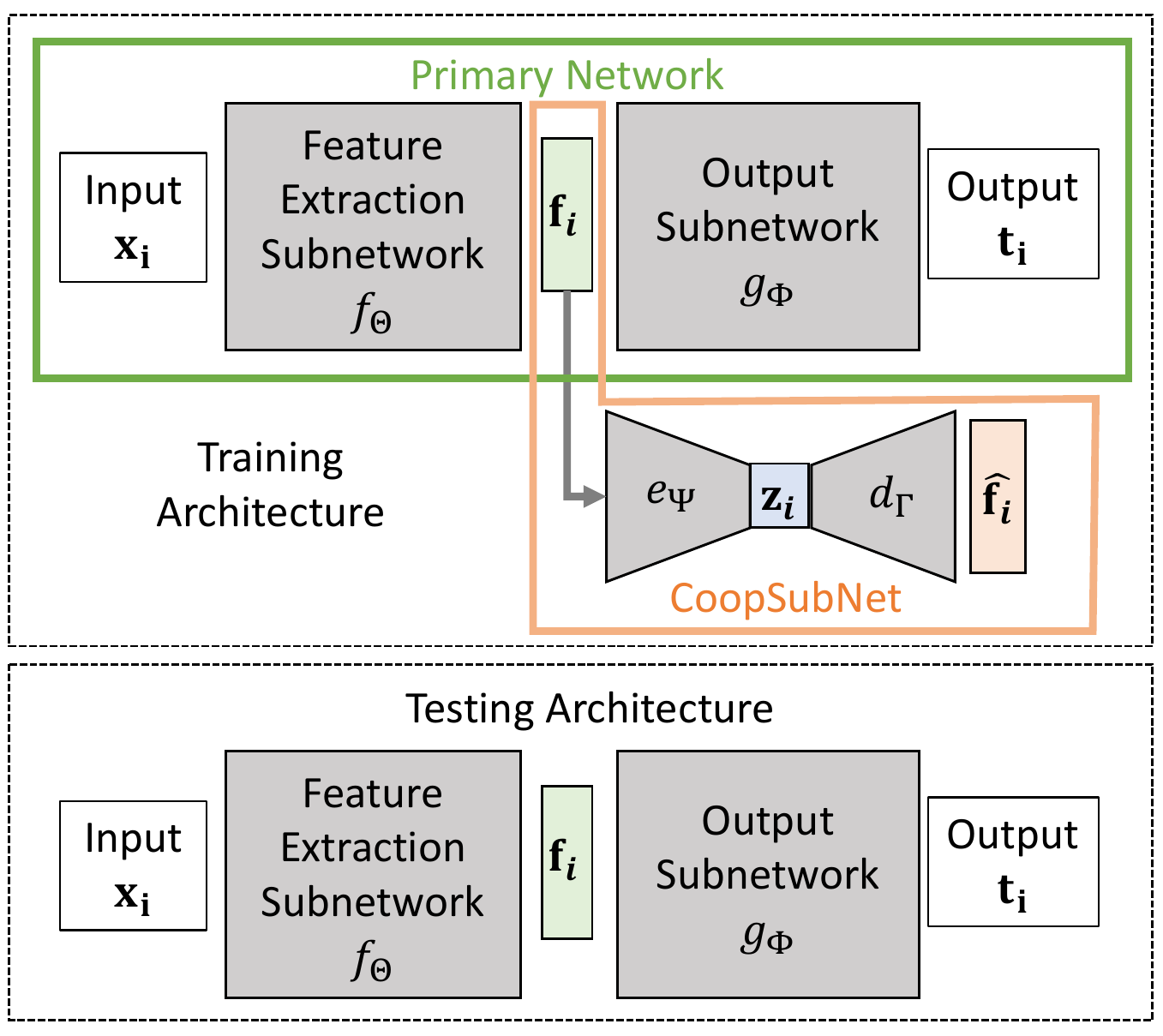}
    \caption{\textbf{CoopSubNet schematic diagram:} A primary network (shown in the green box) comprises two subnetworks, a \textit{feature extraction subnetwork} that learns a task/data-dependent functional mapping between the data space to a latent feature space, and a \textit{output subnetwork} (e.g., classifier, regressor) that maps the feature space to the task-specific output. A CoopSubNet (shown in the orange box), with an auto-encoder architecture, is attached to the feature vector $\mathbf{f}$ of the primary network to produce the reconstruction $\hat{\mathbf{f}}$. The bottleneck of the CoopSubNet, the vector $\mathbf{z}$, defines the intrinsic dimension $L$ of the latent feature manifold.}
    \label{fig:netSchematic}
\end{figure}

\subsection{Cooperating Subnetworks} 

The proposed \emph{cooperating subnetwork} is illustrated in Figure \ref{fig:netSchematic}. The figure shows two interacting networks, the \textit{primary} network (green) that conducts the main learning task (e.g., classification, regression, segmentation), and the \textit{CoopSubNet} as a secondary network (orange) that simultaneously learns a data-driven regularization. 
The specific architecture of the primary network is task-dependent, but can be implicitly divided into two parts: (1) \textit{feature extraction subnetwork} that learns a functional mapping from the data space to a hidden, task-dependent feature space, and (2) an \textit{output subnetwork} that maps the feature space to the final output.
The CoopSubNet can be seen as a regularizer on the feature space, agnostic to the architecture of the primary network. 
To learn the underlying feature distribution or manifold, we train the CoopSubNet as a \emph{bottleneck} auto-encoder.   This architecture alleviates overfitting by forcing or encouraging the learned features to have a statistical structure that is well represented by a low-dimensional, nonlinear model \cite{alain2014regularized}.

Let $\mathcal{D} = \{(\mathbf{x}_1, \mathbf{t}_1), ..., (\mathbf{x}_N, \mathbf{t}_N)\}$ be a training data set with $N-$samples, where $\mathbf{x}_n \in \mathbb{R}^D$ is a training data point that lives in a $D-$dimensional \textit{data} space, and $\mathbf{t}_n \in \mathbb{R}^O$ is the corresponding target label/vector/matrix that lives in an $O-$dimensional \textit{output} space. 
The primary network is parameterized by $\mathcal{P} = \{\Theta, \Phi\}$. $\Theta$ defines the functional mapping $f_\Theta:\mathbb{R}^D \rightarrow \mathbb{R}^F$ that maps a data point $\mathbf{x}$ to a feature vector $\mathbf{f}$ in an $F-$dimensional \textit{feature} space. $\Phi$ defines the mapping $g_\Phi: \mathbb{R}^F \rightarrow \mathbb{R}^O$ to the output space. 
The primary network jointly learns the two mappings using the \textit{task-dependent} primary loss $\mathcal{L}^\mathcal{P} (\Theta, \Phi ; \mathcal{D}) = \frac{1}{N}\sum_n \mathcal{L}^\mathcal{P}_n \left( g_\Phi (f_\Theta ( \mathbf{x_n})) , \mathbf{t}_n \right)$.
The CoopSubNet is attached to the output layer of the feature subnetwork and is parameterized by $\mathcal{C} = \{ \Psi, \Gamma\}$. $\Psi$ represents the parameters of the encoder $e_\Psi: \mathbb{R}^F \rightarrow \mathbb{R}^L$ that maps the primary network's feature space to a latent vector $\mathbf{z}$ in an $L$-dimensional \textit{latent} manifold, where $L \ll F$. $\Gamma$ defines the decoder $d_\Gamma: \mathbb{R}^L \rightarrow \mathbb{R}^F$ that maps latent vectors back to the feature space (produces reconstructed feature $\hat{\mathbf{f}}$).
The CoopSubNet introduces a soft penalty to the primary loss via adding a feature reconstruction loss as a regularizer. We first experimented with a regular auto-encoder loss using $\mathbb{L}_2$ reconstruction loss, but we found there was a problem. In contrast to a stand-alone auto-encoder, the CoopSubNet is trained in conjunction (i.e., at the same time) as the primary network, therefore simply scaling down the variances of the feature space will trivially drive the reconstruction-error loss to zero. This is undesirable, because the same effect can be achieved more easily with a classical data-independent $\mathbb{L}_2$-regularizer. To avoid this trap, we use a \textit{relative} loss, which normalizes the error by the norm of the input features. In terms of equations, the final loss of the \textit{composite} network (i.e., the primary network with a CoopSubNet) is formulated as:
%
%
%
\begin{eqnarray}
\mathcal{L}(\mathcal{P}, \mathcal{C}; \mathcal{D}) &=& \mathcal{L}^\mathcal{P} + \alpha~ \mathcal{L}^\mathcal{C} \nonumber \\
&=& \frac{1}{N}\sum_n \mathcal{L}^\mathcal{P}_n \left( g_\Phi (f_\Theta ( \mathbf{x_n})) , \mathbf{t}_n \right) \nonumber \\
&+& \alpha \frac{1}{N}\sum_n \mathcal{L}^\mathcal{C}_n \left( \mathbf{f}_n, \hat{\mathbf{f}}_n\right)
\label{eq:without-l1}
\end{eqnarray}
\noindent where,
\begin{eqnarray}
 \hat{\mathbf{f}}_n &=& d_\Gamma (e_\Psi ( \mathbf{f}_n)) \\
    \mathcal{L}^\mathcal{C}_n \left( \mathbf{f}_n, \hat{\mathbf{f}}_n\right) &=& \frac{||\mathbf{f}_n - \hat{\mathbf{f}}_n ||^2}{||\mathbf{f}_n||^2}
\end{eqnarray}
In particular, the normalization by $||\mathbf{f}_n||^2$ in the last equation (i.e., the \textit{relative} reconstruction error) ensures that the optimizer does not simply drive $||\mathbf{f}_n||^2$ towards zero.

\subsection{Network Training} 
\label{sub:trainRoutine}
 Training procedure for CoopSubNet consists of two steps. The first step (termed ``\emph{burn-in}'' phase) trains only the primary network, i.e., we set $\alpha = 0$  which corresponds to disabling the CoopSubNet. This is necessary as these burn-in iterations allow for the primary network to learn a preliminary representation of the feature space to bootstrap the training of the CoopSubNet.
In the second step, we jointly train both the primary and the CoopSubNet by setting suitable parameters for $\alpha$. The hyperparameters for the CoopSubNet are the regularization weight $\alpha$, the number of burn-in iterations, the choice for the bottleneck size ($L$), and the choice of where to attach the CoopSubNet, i.e. how to separate the primary network into feature extractor subnetwork and output subnetwork. These hyperparameters are discussed in Section \ref{sec:hyperparameters}.


%

\begin{figure}[!h]
    \centering
    \includegraphics[width=\linewidth]{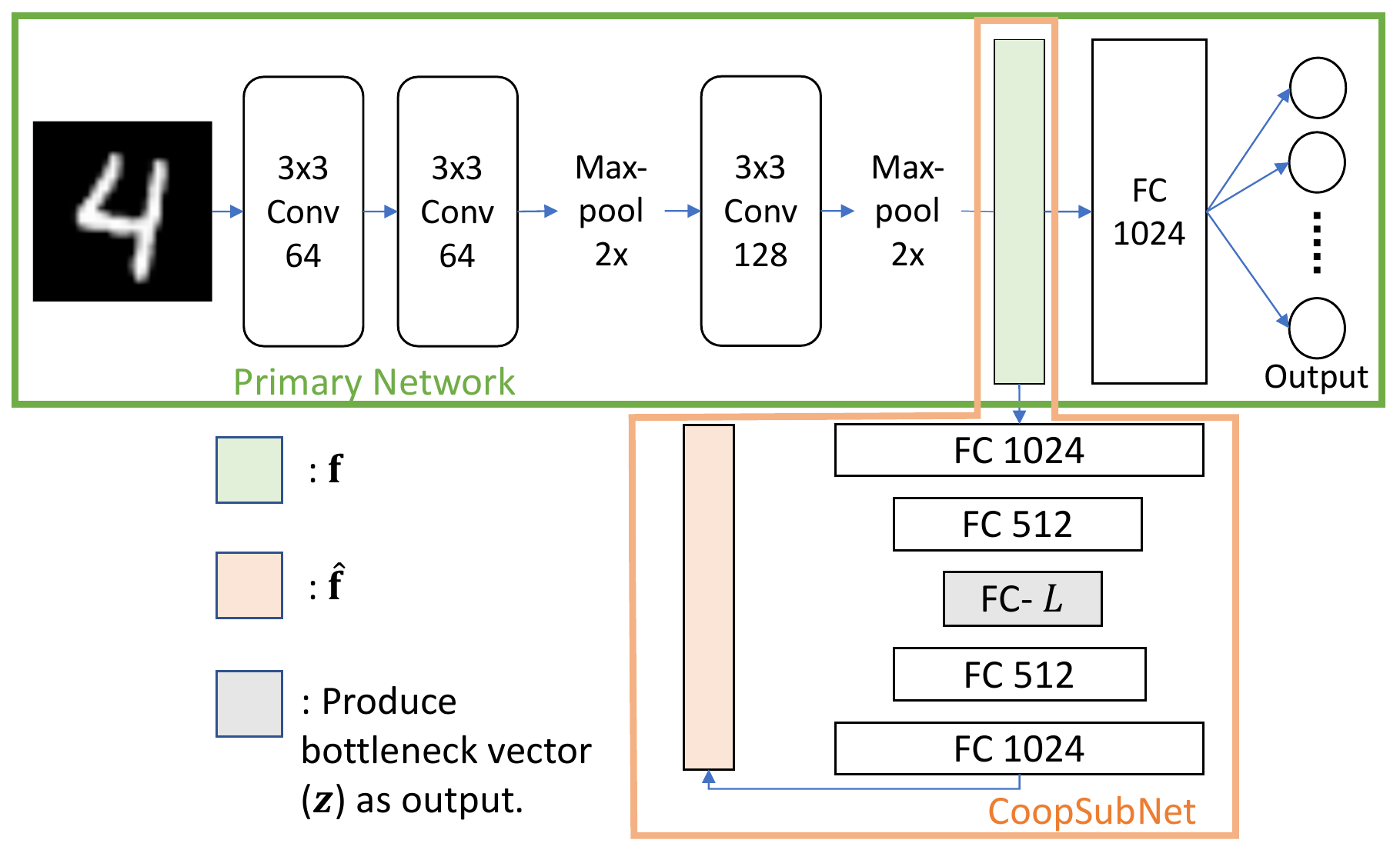}
    \caption{\textbf{MNIST--CoopSubNet/CoopSubNet-L1:} Cooperating subnetwork architecture for MNIST classification. The green box represents the primary network, which uses the cross entropy loss as primary network loss $\mathcal{L}^{\mathcal{P}}$. The CoopSubNet (i.e., cooperating auto-encoder shown in the orange box) is attached to the feature vector $\mathbf{f}$ producing the reconstruction $\hat{\mathbf{f}}$. }
    \label{fig:mnistArch}
\end{figure}

\section{Results}
\label{sec:results}
\begin{table*}[!h]
\centering
\caption{\textbf{Reduced MNIST classification accuracy results:}  CoopSubNet and HardCon bottlenecks are set to $L = 16, 64, 256, 512,$ and $512$ for $0.5\%, 1\%, 10\%, 50\%,$ and $ 100\%$ training data, respectively.}
\vspace{4pt}
\begin{tabular}{|l||l|l|l|l|l|l|}
\hline
 Reduced Data  & 0.5 \%   & 1 \%  & 10 \% & 50 \% & 100 \% \\ \hline
 Baseline   & 90.2 $\pm$ 0.41  & 93.2 $\pm$ 0.28  &  97.9 $\pm$ 0.05  & 98.9 $\pm$ 0.09  & 99.3 $\pm$ 0.02     \\ \hline
                        CoopSubNet    & \textbf{93.7 $\pm$ 0.42}  & \textbf{95.8 $\pm$ 0.27}  & \textbf{98.2 $\pm$ 0.06} & 98.9 $\pm$ 0.07  & \textbf{99.4 $\pm$ 0.00}   \\ \hline 
                        Dropout        & 90.5 $\pm$ 0.02 & 93.5 $\pm$ 0.01 & 97.8 $\pm$ 0.00  & \textbf{99.0 $\pm$ 0.00} & 99.2 $\pm$ 0.00   \\ \hline
                        L2-reg         & 90.4 $\pm$ 0.32 & 94.1 $\pm$ 0.21 & 97.4 $\pm$ 0.03 & 98.0 $\pm$ 0.01 & 99.0 $\pm$ 0.00    \\ \hline
                        HardCon    &  90.5 $\pm$ 0.38 & 93.5 $\pm$ 0.30  & 98.1 $\pm$ 0.02 & 98.9 $\pm$ 0.00   &  99.2 $\pm$ 0.00  
                           \\ \hline
\end{tabular}
\label{tab:MCresults}
\end{table*}

\begin{figure}[!h]
    \centering
    \includegraphics[width=\linewidth]{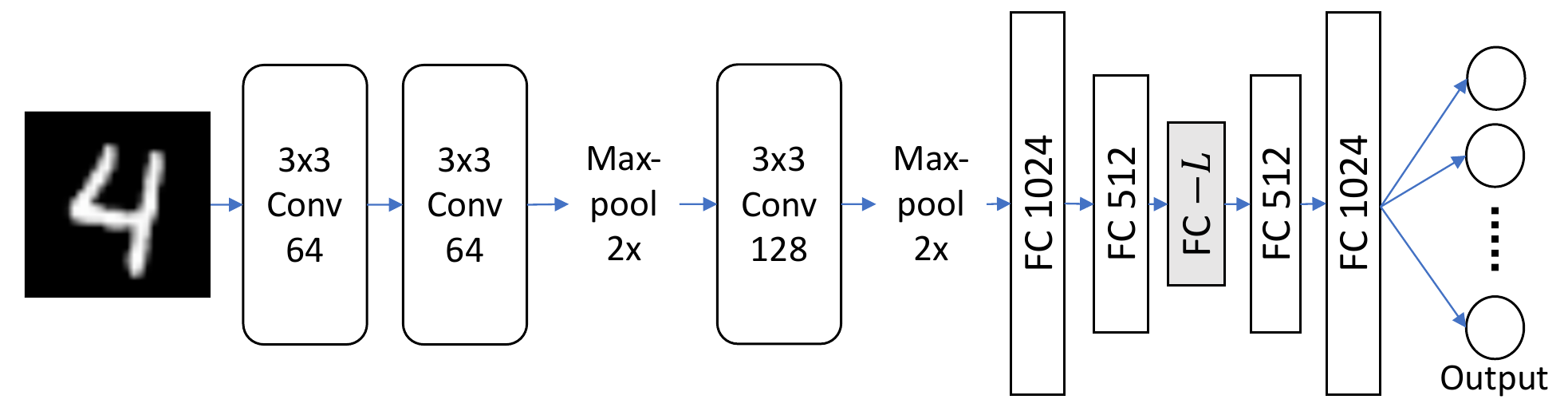}
    \caption{\textbf{MNIST--HardCon:} Hard constraint (HardCon) network architecture variant for MNIST classification. The dimension of the latent space $L$ acts as the hard constraint. We use the same $L$ for CoopSubNet and HardCon versions in all of our comparisons.}
    \label{fig:mnistArch-hardcon}
\end{figure}

The proposed CoopSubNet framework is effective in improving supervised training accuracy in situations with limited data budget. We demonstrate this on two tasks, \textit{classification} and \textit{regression}. 
In our experiments we compare with the following methods: 
(1) \textit{Baseline} is a task-dependent architecture that is trained without any regularization. 
(2) \textit{CoopSubNet} is an auto-encoder with a predefined bottleneck attached to the feature layer of the primary network. The composite network is trained based on the loss in Eq.~\ref{eq:without-l1}.
%
%
(3) \textit{Dropout} is a task-dependent architecture that include dropout layers \cite{srivastava2014dropout}.
(4) \textit{L2-reg} uses $\mathbb{L}_2$-norm on the primary network parameters as a generic weight decay regularizer \cite{lang1990dimensionality,krogh1992simple}.
(5) \textit{HardCon} enforces the manifold assumption as a hard constraint by restricting the dimensionality of the bottleneck directly in the primary network, i.e., force features to lie on a low-dimensional manifold. Figure \ref{fig:mnistArch-hardcon} illustrates adding such a hard bottleneck to the primary network for MNIST classification. We evaluate this hard-constraint variant for each of our experiments to provide insights about how enforcing a hard architectural constraint performs in comparison to a soft penalty as enforced by the CoopSubNet. 
In all experiments we use the same bottleneck dimension, $L$, for both CoopSubNet and HardCon. 
We initialize all the networks considered for comparison (including CoopSubNet variants) with Xavier's initialization \cite{Glorot10understandingthe}, use batch normalization on all the convolution layers \cite{sergey2015BatchNormalization}, and train using the Adam optimizer \cite{kingma2014adam}. 

\subsection{Reduced MNIST}
\label{sub:RedMNIST}
We begin our experimental evaluation with a classical image classification problem. Specifically, we use a reduced version of the MNIST dataset (the full dataset contains  
60,000 training and 10,000 testing images of handwritten digits). 
%
The ``reductions'' consists in randomly selecting a subset of the original MNIST data, simulating conditions with limited data budgets. The proposed cooperating subnetworks architecture is described in Figure \ref{fig:mnistArch} and we use the training procedure described in section \ref{sub:trainRoutine}. We demonstrate the efficacy of CoopSubNet trained on 0.5\%, 1\%, 10\% and 50\% of the total training data size, and evaluate the testing accuracy on the entire testing data. 
%
%
%
The bottlenecks for the CoopSubNet were chosen using cross-validation, and are set to $L = 16, 64, 256, 512,$ and $512$ for $0.5\%, 1\%, 10\%, 50\%,$ and $ 100\%$ training data, respectively.
The results are summarized in Table \ref{tab:MCresults} using the  \% of classification accuracy on the test set. Each experiment was repeated with 5 randomly drawn training subsets; we report the means and standard deviations. 
%
%
In all of the experiments, the CoopSubNet learns to reconstruct well, with  relative reconstruction loss  approximately 1\% (on entire testing data).  
The proposed architecture (CoopSubNet) shows significant improvement over other methods especially with limited amounts of training data, in particular $0.5\%$ and $1\%$ of the original training set. With more training data, the benefits of using CoopSubNet diminish, however, the results with CoopSubNet are still on par or even slightly better than with other methods. 

Let us discuss the comparison of CoopSubNet and the hard-constraint variant (HardCon), which forces the features to lie on a low-dimensional manifold exactly. The HardCon version corresponds to changes of network architecture, while CoopSubNet correspond to  regularization. The results in Table \ref{tab:MCresults} indicate the soft constraint in CoopSubNet leads to better generalization than the hard constraint. We further study this in Table \ref{tab:hardconBreak}, using networks trained on $1\%$ of the training data with latent space dimensions set to: 1, 4 and 16. We can see that if the dimension of the bottleneck is set too low, HardCon's accuracy reduces to random chance, while CoopSubNet still perform well. At $L = 16$, the HardCon version starts to work, but CoopSubNet still produce better accuracy.


\begin{table}[]
\centering
\caption{\textbf{HardCon versus CoopSubNet:} Results on reduced MNIST data trained with 1\% of the training data using three different bottleneck/latent dimensionality. With a very constrained bottleneck, HardCon does not learn the underlying feature manifold and drastically affect the classification performance of the primary network.}
\vspace{4pt}
\begin{tabular}{|l|l|l|}
\hline
Bottleneck ($L$) & HardCon & CoopSubNet \\ \hline
1 & 9.8     & \textbf{94.7}               \\ \hline
4 & 9.8     & \textbf{94.9}           \\ \hline
16 & 93.1    & \textbf{95.1}            \\ \hline
\end{tabular}
\label{tab:hardconBreak}
\end{table}

\begin{figure*}[!h]
    \centering
    \includegraphics[width=\linewidth]{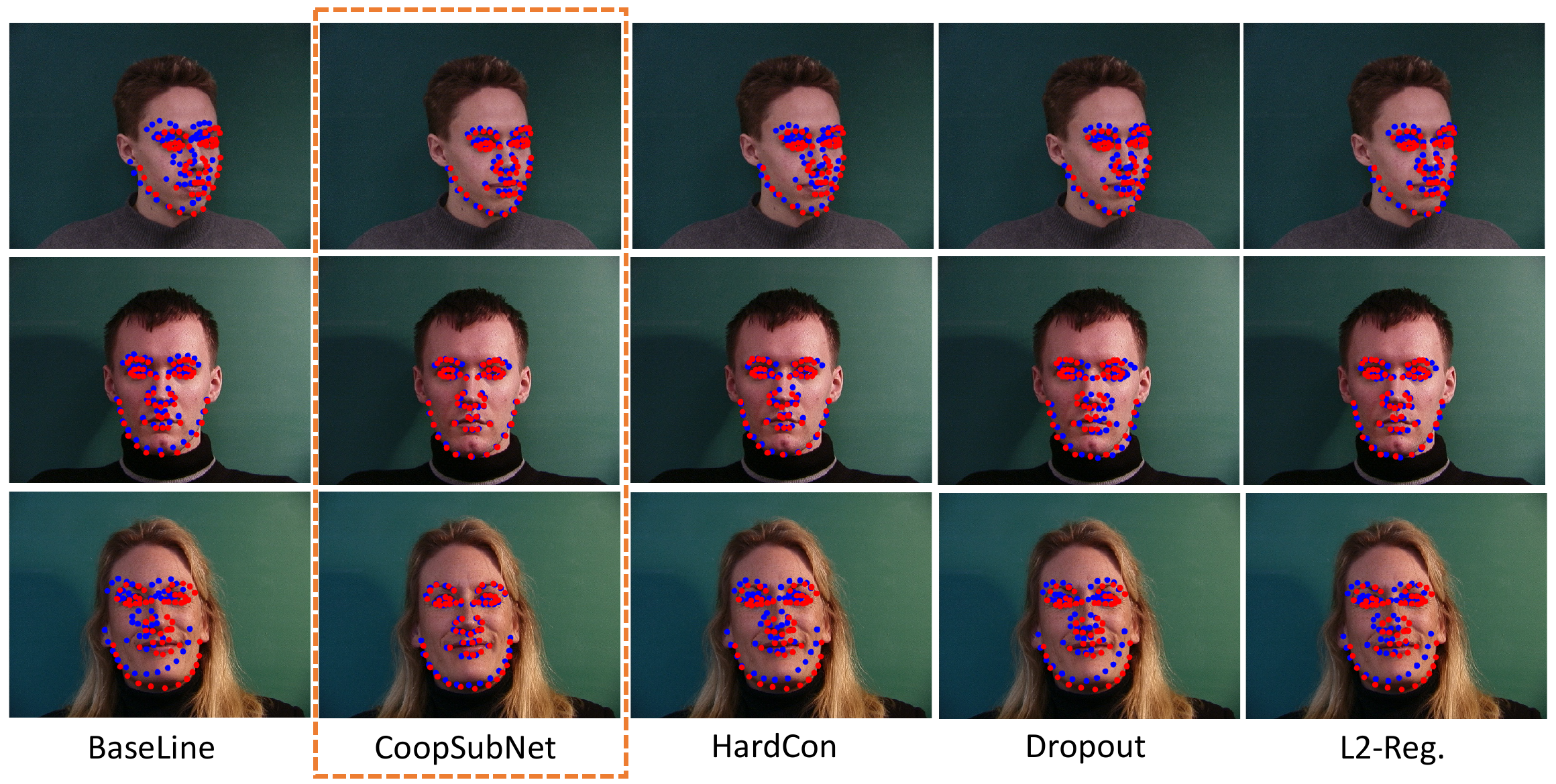}
    \caption{\textbf{Landmark detection qualitative results:} The landmark detection results on three different test images. The points marked in \textcolor{red}{red} are the ground truth landmarks, and the points in \textcolor{blue}{blue} are the ones predicted by different networks.}
    \label{fig:landmarkOutput}
\end{figure*}

\begin{figure}[!h]
    \centering
    \includegraphics[width=\linewidth]{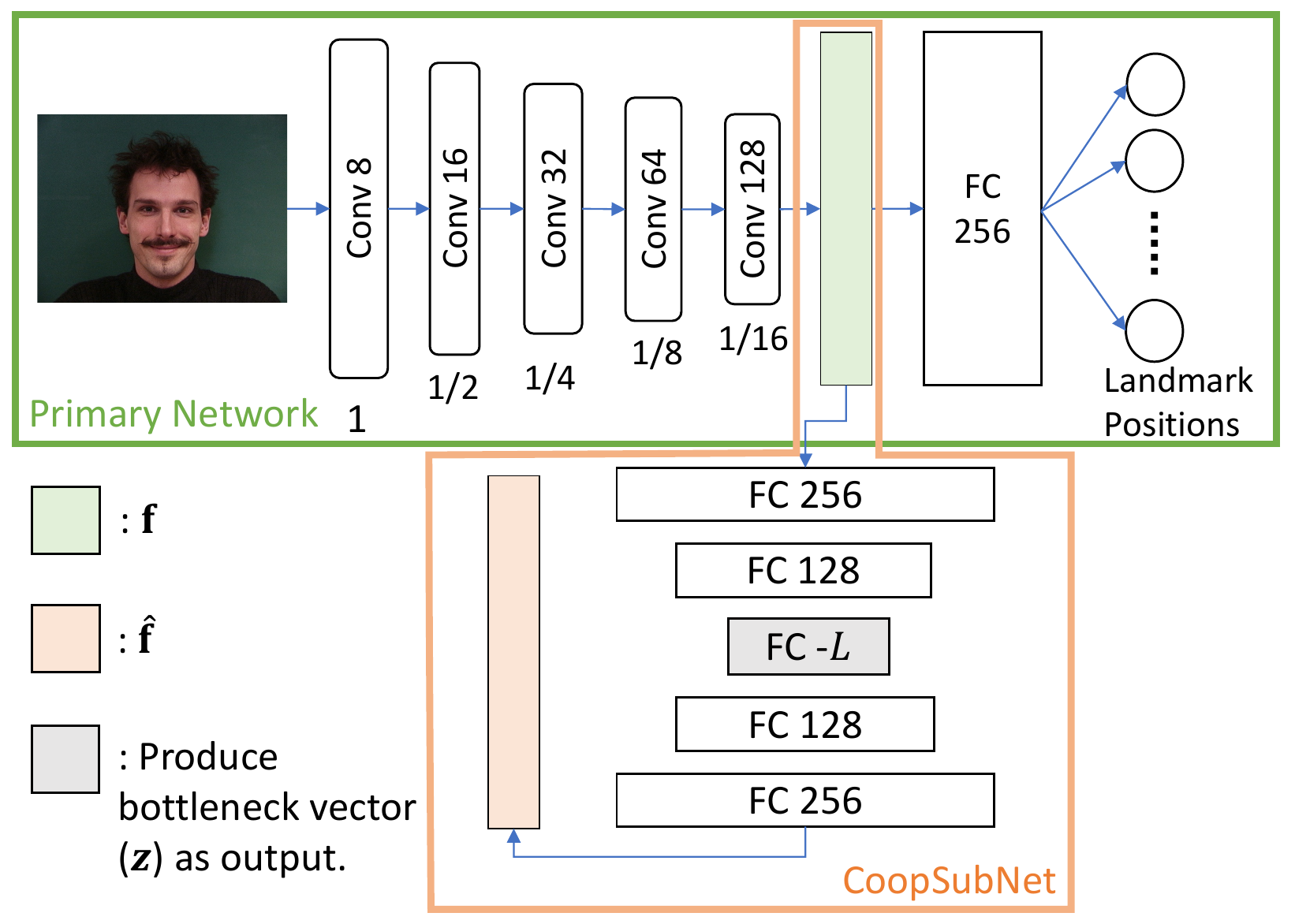}
    \caption{\textbf{Landmark detection -- CoopSubNet/CoopSubNet-L1:} Cooperating subnetwork architecture for the landmark detection task. The primary loss is an $\mathbb{L}_2$ difference between network estimated landmarks positions and the ground truth ones.}
    \label{fig:archLandmarks}
\end{figure}

\begin{figure}
    \centering
    \includegraphics[width=\linewidth]{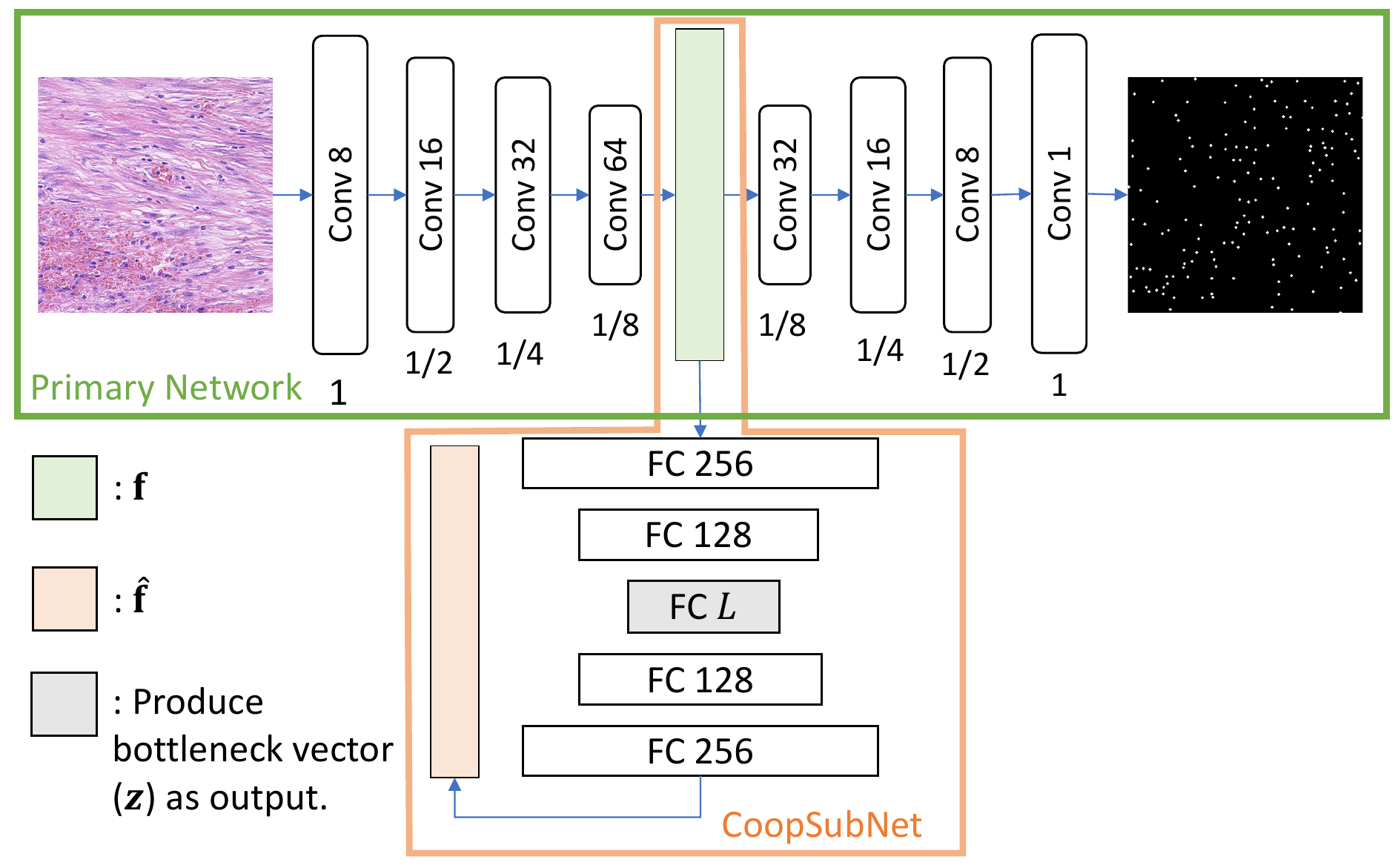}
    \caption{\textbf{Nuclei estimation -- CoopSubNet/CoopSubNet-L1:} Cooperating subnetwork architecture for cell nuclei estimation. The primary network predicts a probability map, and the primary loss is an $\mathbb{L}_2$ difference between the estimated probability map and the ground truth nuclei map.}
    \label{fig:nucleiArch}
\end{figure}

\begin{figure*}[!h]
    \centering
    \includegraphics[width=\linewidth]{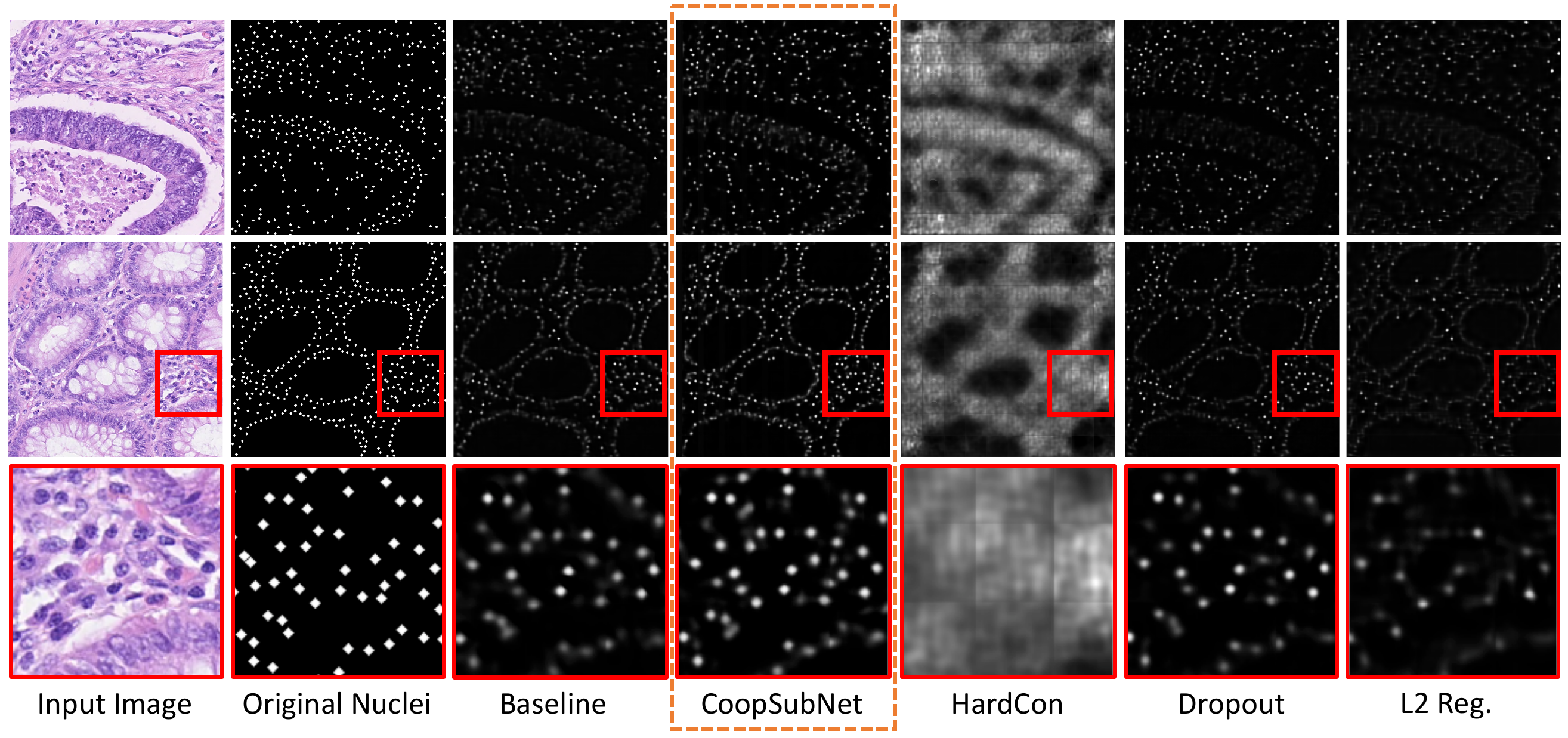}
    \caption{\textbf{Nuclei estimation results} on two different test samples. A zoom-in view of the red highlighted region in the second row is shown in the third row.}
    \label{fig:nucleioutput}
\end{figure*}

\subsection{Facial Landmark Detection}
\label{sub:landmark}

Detecting facial landmarks in images is an example of a regression problem, belonging to the broader class of image-to-landmark problems studied in computer vision.
%
Previous work shows that facial landmarks lie on a low-dimensional manifold \cite{chang2006manifoldface}, justifying the manifold assumption in CoopSubNet.
%
In our experiments we use the IMM facial landmark dataset \cite{stegmann2004IMM}, which contains 240 facial images of 24 different people with different head poses and illumination conditions. Each $480 \times 640$ image is annotated with 58 landmarks with correspondences. Because the annotation process is time consuming, the size of the dataset is relatively small for effective training of convolutional neural networks, making this task a perfect match for CoopSubNet.
%
We split the IMM dataset into a training set (75\%) and a test set (25\%).
The network architecture is described in Figure \ref{fig:archLandmarks}. 
%
Results on the test set are computed in terms of squared errors between the predicted and ground truth facial landmarks and are enumerated in Table \ref{tab:landmarkResults}. Each experiment was performed 5 times with a different sampling of the training and testing data each time (i.e., random splits). The mean results are reported along with the standard deviations. 
We can see that the CoopSubNet outperforms all other methods by a significant margin. The higher accuracy of landmark prediction can be seen even by the naked eye, see Figure \ref{fig:landmarkOutput}. 
The latent space dimension ($L$) used for CoopSubNet for these results was 32. As in the case with MNIST experiments, we ensure that the CoopSubNet has a relative reconstruction loss of approximately 1\%(on entire testing data).



\begin{table}[!h]
\centering
\caption{\textbf{Landmark detection quantitative results:} CoopSubNet and HardCon bottlenecks are set to $L = 32$.}
\vspace{4pt}
\begin{tabular}{|l|l|}
\hline
Method      & Landmark Error  \\ \hline
Baseline    & 6.86 $\pm$ 1.23   \\ \hline
CoopSubNet & \textbf{3.03 $\pm$ 0.83}   \\ \hline
Dropout     & 6.31 $\pm$ 0.34   \\ \hline
L2-reg      & 6.41 $\pm$ 0.66 \\ \hline
HardCon & 5.89 $\pm$ 0.65   \\ \hline
\end{tabular}
\label{tab:landmarkResults}
\end{table}

\subsection{Cell Nuclei Probability Estimation}
\label{sub:nuclei}



\begin{table*}[!h]
\centering
\caption{\textbf{Nuclei Estimation Results.} The bottlenecks for HardCon and CoopSubNet are set to $L = 64$. We report the reconstruction error (L2 Error), segmentation accuracy (Dice coefficient), nuclei detection accuracy (precision, recall and F1-score).}
\vspace{4pt}
\begin{tabular}{|l|l|l|l|l|l|}
\hline
Method       & L2 Error & Dice  & Precision & Recall & F1-Score \\ \hline
Baseline     &  0.0225 $\pm$ 0.005 & 0.548 $\pm$ 0.012 & 0.853 $\pm$ 0.005 & 0.662 $\pm$ 0.013 & 0.741 $\pm$ 0.008\\ \hline
CoopSubNet     & \textbf{0.0192 $\pm$ 0.002} & \textbf{0.564 $\pm$ 0.020} & 0.846 $\pm$ 0.007 & \textbf{0.735 $\pm$ 0.022} & \textbf{0.794 $\pm$ 0.011} \\ \hline
Dropout       & 0.0205 $\pm$ 0.001 & 0.528 $\pm$ 0.002 & \textbf{0.878 $\pm$ 0.002} & 0.658 $\pm$ 0.006 & 0.747  $\pm$ 0.005 \\ \hline
L2-Reg        & 0.0233 $\pm$ 0.000 & 0.418 $\pm$ 0.032 & 0.769 $\pm$ 0.010 & 0.528 $\pm$ 0.025 & 0.621 $\pm$ 0.025   \\ \hline
HardCon       & 0.0291 $\pm$ 0.007 & 0.107 $\pm$ 0.004 & 0.064 $\pm$ 0.001 & 0.011 $\pm$ 0.002 & 0.013 $\pm$ 0.002  \\ \hline
\end{tabular}

\label{tab:nuclieResults}
\end{table*}
Cell nuclei detection is an image segmentation problem, which can be formulated as an image-to-image regression task, where the output image is a probability map (pixels represent probabilities of corresponding to a cell nucleus).
%
This approach has been studied in recent work \cite{hofner2018simpleNuclei,alom2018DCNNNuclei}. 
Here, we use a public dataset from the Tissue Image Analytics Lab at Warwick, UK, as described in \cite{sirinukunwattana2016locality}. This dataset contains one hundred $500 \times 500$ H\&E stained histology images of colorectal adenocarcinomas, with more than 25,000 nuclei centers marked for detection tasks. We split this dataset into 75\% training and 25\% testing data. The CoopSubNet architecture for this task is shown in Figure \ref{fig:nucleiArch}. We train this network on image patches of $250 \times 250$ pixels with a stride of 50 in both dimensions, which results in 2700 images for training and 900 images for testing. The associated binary label map has a single pixel highlighted at the nuclei center, which makes the learning process much harder due to unbalanced class labels. To improve this, we perform a single morphological dilation iteration using a $3 \times 3$ square structuring unit. For CoopSubNet and reference methods we use $\mathbb{L}_2$ difference between the estimated probability map and the ground truth nuclei map as the primary loss. We set the bottleneck dimension to 64 for CoopSubNet and HardCon. The results are demonstrated visually in Figure \ref{fig:nucleioutput}, and quantitatively in Table \ref{tab:nuclieResults}. The relative reconstruction loss of CoopSubNet on testing data is less than 2\%. The proposed data-driven regularization mechanism shows qualitative (F1-Score) and quantitative improvement over other regularization methods. In particular, the recall improves significantly, implying a better detection of nuclei and fewer false negatives. This can also be visually observed in the zoomed in region of the second sample shown in third row of Figure \ref{fig:nucleioutput}. We can notice the HardCon results are quite blurry because the bottleneck dimension of 64 is insufficient for precise localization of the nuclei positions.


\vspace{0.1in}
\noindent\textbf{Quantitative metrics:} We use different evaluation metrics to quantify the performance of the nuclei detection task. First, we use an Euclidean norm distance between the output probability map and the ground truth labels ($G_L$). To compute detection-based metrics, we binarize the probability map using Otsu thresholding \cite{kurita1992maximum}, a clustering-based image thresholding, to obtain a binary output label map ($O_L$).
%
The segmentation accuracy is evaluated using the Dice coefficient \cite{taha2015metrics} that quantify the spatial overlap between the $G_L$ and $O_L$. For precision, recall and F1 scores, we employ a simple algorithm  that captures the true essence of the number of nuclei detected.
%
Starting with the $O_L$, we apply connected components and compute the position of the center of mass for each connected component. These positions represent each detected nucleus. To find the true and false positives, we query $G_L$ at each of the positions, and if it corresponds to a \emph{unique} nucleus then it is counted as a true positive (TP); otherwise, it is a false positive (FP). To compute the false negatives (FN), we count the number of nuclei in $G_L$ that were not queried at all. Using these three values, TP, FP and FN, we can compute the metrics, as follows: $\mathrm{precision} = \frac{TP}{TP + FP}$, $\mathrm{recall} = \frac{TP}{TP + FN}$ and, $F_1 = 2\frac{\mathrm{precision} \times \mathrm{recall}}{\mathrm{precision} + \mathrm{recall}}$.

\subsection{Hyperparameter Selection}
\label{sec:hyperparameters}

CoopSubNet has four major hyperparameters. Here, we discuss the strategies for their tuning. 

\noindent\textbf{Bottleneck Selection: } The dimension $L$ of the bottleneck layer of the secondary network in CoopSubNet is an important hyperparameter that affects the training dynamics of the composite network. A smaller latent dimensionality magnifies the regularization effect, but a too small $L$ could derail learning the intrinsic structure of the underlying feature manifold. From our experiments, we found that based on the dataset and network architecture there is a sweet spot for $L$ that delivers the best performance. 
Alternatively, instead of tuning directly the dimension $L$ of the bottleneck, we tried to make the bottleneck conservatively large and instead impose an $\mathbb{L}_1$ penalty on the latent vector $\mathbf{z}_n$ (Figure \ref{fig:netSchematic}) in addition to the composite network loss (Eq \ref{eq:without-l1}). 
In our experiments with this alternative approach, we found that this sparse penalty on a large-enough bottleneck leads to similar results as CoopSubNet.
However, the direct CoopSubNet approach is easier to use.

\noindent\textbf{Contribution of the CoopSubNet loss:} The second hyperparameter is the weight of the CoopSubNet loss in the composite network loss ($\alpha$). In our experiments, we found that as long as the CoopSubNet loss has the same order of magnitude as the primary loss, the performance of the regularized primary network is not sensitive on the specific value of $\alpha$. 
%

\noindent\textbf{Burn-in iterations:} We found that results are not particularly sensitive to the initialization phase (``burn-in'').  In all experiments, we set the burn-in to $5\%$ of the total epochs. 

\noindent\textbf{Feature space:} We have a choice of which layer of the primary network we attach the CoopSubNet to (Figure \ref{fig:netSchematic}). We experimented with different options in the MNIST architecture (Figure \ref{fig:mnistArch}). We found that if we attach the CoopSubNet too close to the input (after the first or second convolutional layer), the CoopSubNet does not regularize as well. This can be attributed to the fact that the features are not yet defined enough to have an underlying low-dimensional  representation. On the other hand, the performance of CoopSubNet only changes marginally when attached to layers close to the output layer of the primary network.

\section{Conclusions and Future Work}
\label{sec:discussion}

This paper introduces a novel neural network architectural framework: a secondary cooperating subnetwork (CoopSubNet) acting as a data-dependent regularizer. The CoopSubNet is a flexible framework and can be used alongside any type of primary architecture. It is trained jointly with the primary network to enforce a soft constraint on the intermediate features ensuring they lie close to a low-dimensional manifold. The experimental results show that CoopSubNet is an effective regularization technique as compared to standard regularization methods in problems entailing limited data budgets.

In terms of future research, other approaches to data-driven regularization could be studied. Our experiments provide promising results with the $\mathbb{L}_1$ penalty on bottleneck vector of CoopSubNet (discussed in Section \ref{sec:hyperparameters}). This motivates exploration of other sparse penalties such as the automatic relevance determination (ARD) prior \cite{tipping2001sparse} or denoising autoencoders \cite{vincent2010stacked}. Other future research directions include exploring the performance of CoopSubNet in a system of networks such as GANs, or extending CoopSubNet to a semi-supervised learning framework, where the training data are only partially labeled.

{\small
\bibliographystyle{ieeetr}
\bibliography{main}
}

\end{document}